\documentclass[letterpaper, 10 pt, conference]{ieeeconf}  % Comment this line out if you need a4paper

\IEEEoverridecommandlockouts                              % This command is only needed if 

\usepackage{graphics} % for pdf, bitmapped graphics files
\usepackage{epsfig} % for postscript graphics files
\usepackage{mathptmx} % assumes new font selection scheme installed
\usepackage{times} % assumes new font selection scheme installed
\usepackage{amsmath} % assumes amsmath package installed
\usepackage{amssymb}  % assumes amsmath package installed

\usepackage{algorithm}
\usepackage{algorithmic}

\usepackage{booktabs}
\usepackage{multirow}
\usepackage{threeparttable}
\usepackage{comment}
\usepackage{graphicx}
\usepackage{threeparttable}
\usepackage{xcolor}
\usepackage[normalem]{ulem}

\usepackage{mathrsfs}
\usepackage{dutchcal}

\definecolor{revisiongreen}{RGB}{0,128,0}
\definecolor{draftnoteviolet}{RGB}{128,0,160}
\definecolor{revisionnoteorange}{RGB}{190,90,0}

\title{\LARGE \bf
Task-Oriented Active Learning of Residual Dynamics\\
for Model Predictive Path Integral Control
}

\author{Nobuaki Aoki$^{1}$, Hojin Lee$^{2}$, Stefan Sosnowski$^{1}$, Sandra Hirche$^{1}$
\thanks{$^{1}$Nobuaki Aoki, Stefan Sosnowski and Sandra Hirche are with Chair of Information-oriented Control (ITR), Department of Electrical and Computer Engineering, Technical University of Munich, Munich, Germany
        {\tt\small $\{$nobuaki.aoki; sosnowski; hirche$\}$@tum.de}}
\thanks{$^{2}$Hojin Lee is with the Munich Institute of Robotics and Machine Intelligence (MIRMI), Technical University of Munich, Munich, Germany. {\tt\small hojin.lee@tum.de}
}
}

\begin{document}

\maketitle
\thispagestyle{empty}
\pagestyle{empty}

%%%%%%%%%%%%%%%%%%%%%%%%%%%%%%%%%%%%%%%%%%%%%%%%%%%%%%%%%%%%%%%%%%%%%%%%%%%%%%%%

\begin{abstract}
Online residual learning can reduce model mismatch in predictive control, but passive data collection may fail to adequately cover states that become important later in the task.
Task-agnostic active learning targets uncertain or informative regions, but information acquired in such regions does not necessarily improve task performance.
This paper introduces Task-Oriented Information Acquisition (ToIA), an active-learning criterion for model predictive path integral control (MPPI) with online Gaussian process (GP) residual learning.
For each sampled control sequence, ToIA estimates how much an observation obtained early in the rollout would reduce predictive uncertainty at later states on the same rollout, and weights this reduction by the rollout's relevance to the task.
The score is evaluated over the existing MPPI rollout batch without sampling future observations or re-optimizing control under hypothetical posterior updates.
In simulated off-road navigation across held-out maps with heterogeneous terrain, ToIA improved the goal-reaching success rate over passive GP learning by 19.3 and 27.4
percentage points and outperformed task-agnostic active-learning baselines across dense and sparse online-learning intervals.
An ablation study indicates that task relevance is particularly important under sparse model updates.
The implementation supports online control at 20 Hz on an NVIDIA RTX 2080 Ti.

\end{abstract}

\section{Introduction}

Model-based predictive control depends on a model that is sufficiently accurate to determine effective control actions.
However, unmodeled and difficult-to-characterize environmental effects can produce substantial dynamics mismatch, as in terrain-dependent
ground-vehicle dynamics \cite{levy_meta_offroad_rss25,trivedi_stochastic_mppi_icra25},
wind-induced aerodynamic effects on aircraft \cite{neuralflySciRobo2022},
and flow-dependent hydrodynamics on underwater vehicles
\cite{Amer2025TCSTunderwater}.
Online model adaptation and residual learning address this mismatch by updating the prediction model from data collected during closed-loop
operation.

Recent learning-based predictive-control methods have demonstrated the benefit of online dynamics adaptation using Gaussian processes (GPs) \cite{Amer2025TCSTunderwater} or neural networks \cite{saviolo2023active, jiahao2023online}. 
In passive online learning, however, the model is updated only from data encountered during closed-loop operation, without directing data acquisition toward states that are likely to be encountered later in the task.
Consequently, such states may remain poorly represented in the data.
Active dynamics-learning methods instead seek uncertain or informative state--action regions to improve sample efficiency \cite{buissonfenet2020jointIG}.
Yet reducing model uncertainty does not necessarily improve control performance.

Prediction-oriented active learning makes this distinction
explicit.
Expected predictive information gain (EPIG) values candidate observations
through their effect on predictions under a target input distribution
\cite{bickfordsmith2023epig}, while targeted active learning for Bayesian
decision-making values information according to its effect on a downstream
decision \cite{filstroff2024targeted}.
These approaches demonstrate the general principle that the value of information depends on where the resulting predictions or decisions are used, rather than on global model uncertainty alone.
They do not, however, directly address closed-loop dynamics learning, where the control action determines both which observation becomes available and which future operating points are subsequently encountered.

In dynamical systems, this coupling appears through the dual effect of control: actions affect both the physical state and the information available for future decisions.
Active and dual MPC methods therefore account for excitation or uncertainty reduction and, in some cases, anticipate how the resulting observations update the model while balancing these benefits against immediate control performance
\cite{parsi2020active,arcari_dual_mpc_l4dc,arcari2020approximate}.
Localized active learning anticipates future information gain within MPC over a predetermined region of interest \cite{capone2020localized};
trajectory-level active GP dynamics learning selects informative reachable state--action sequences \cite{buissonfenet2020jointIG}; and Bayesian MPC uses posterior sampling to couple model exploration with predictive control \cite{wabersich2020bayesian}.
These methods, however, do not explicitly value the acquired information according to its relevance to downstream task performance.

Related ideas also appear in reinforcement learning, where Bayesian exploration accounts for uncertainty in unknown dynamics \cite{klenske2016dual}.
Trajectory Information Planning, for example, plans exploratory actions to maximize information about the task-optimal trajectory \cite{mehta_NEURIPS2022_tip}, while task-informed exploration policies can be learned from the sensitivity of task execution to uncertain physical properties and followed by a transition to task execution
\cite{aoyama_task_informed_corl25}.
Such approaches tie information acquisition more directly to downstream task value, but can require posterior model sampling, repeated task optimization,
or a separately learned exploration policy.
Bayesian dual control, in turn, accounts for how current actions affect future observations, posterior beliefs, and subsequent control decisions.
Related work on learning-based control has also considered how future online model updates affect closed-loop performance \cite{capone2020anticipating}.
Exact formulations require stochastic dynamic programming over the belief state and are generally computationally intractable for online control \cite{arcari_dual_mpc_l4dc,klenske2016dual}.
This leaves a practical gap between task-agnostic uncertainty seeking and exact dual control when task execution and model learning must occur simultaneously in receding-horizon control.

To address this issue, we propose Task-Oriented Information Acquisition (ToIA), a tractable active-learning mechanism within a model predictive path integral (MPPI) framework
\cite{williams_mppi_2017icra,williams2017model}, using Gaussian process
(GP) regression \cite{williams2006gaussian} for online residual learning.
For each sampled rollout, ToIA considers observations that would become available at early steps and evaluates how much they would reduce predictive uncertainty at later states on the same rollout.
This predictive value is then combined with a task-relevance weight derived from the rollout cost.
The acquisition score is incorporated into the rollout cost to bias the MPPI weighting toward rollouts that provide useful learning opportunities while remaining relevant to the current task.

We evaluate whether ToIA improves closed-loop task performance relative to passive residual learning and task-agnostic active-learning baselines.
The contributions are:
\begin{enumerate}
    \item A predictive acquisition formulation that evaluates how observations obtained at early steps of a rollout would reduce uncertainty at later states on the same rollout, while accounting for the rollout's relevance to the task.
    \item An analytic GP formulation that evaluates the acquisition within the existing MPPI rollout batch without sampling future observations or re-optimizing control under hypothetical posterior updates, supporting 20-Hz online execution.
    \item An evaluation on held-out heterogeneous terrain against passive and task-agnostic active-learning baselines, including matched-condition comparisons and performance across terrain types that induce different residual dynamics.
\end{enumerate}

The remainder of this paper is organized as follows.
Section~\ref{sec:problem_statement} formulates online residual learning and the MPPI control problem.
Section~\ref{sec:toia} introduces ToIA and its predictive acquisition criterion.
Section~\ref{sec:experimental_design} describes the experimental design, and
Section~\ref{sec:results} presents the closed-loop evaluation and mechanism analysis.
Section~\ref{sec:discussion} discusses the implications and limitations of the proposed approach,
and Section~\ref{sec:conclusion} concludes the paper.

\textit{Notation:}
Let $\mathbb R$ denote the real numbers, $I$ the identity matrix, and $\mathbf{1}\{\cdot\}$ the indicator function.
Bold lowercase letters denote vectors.
The index $t$ denotes closed-loop time, $k$ indexes sampled MPPI rollouts, $h$ indexes prediction steps within a rollout, and $j$ indexes scalar components of the residual model.
For a scalar GP, $\mu(\cdot)$ and $\Sigma(\cdot,\cdot)$ denote its posterior mean and covariance, respectively; subscripts identify the
corresponding GP.
For a finite set $\mathcal S$, $|\mathcal S|$ denotes its cardinality.

\section{Problem Formulation}
\label{sec:problem_statement}

This section formulates the residual dynamics, online learning model,
and receding-horizon controller used in this paper.
The specific vehicle model and terrain-dependent residuals used in the
experiments are given in Sec.~\ref{sec:experimental_design}.

\subsection{Residual Dynamics and Online Learning}
\label{subsec:prob_form_residual_learning}

We consider a system with known nominal dynamics and an unknown nonlinear residual.
In robotic systems, such residuals can capture effects that are difficult to model a priori, including tire--terrain interaction, aerodynamic disturbances, and hydrodynamic forces.

Let $n_r$ denote the number of residual components.
For each component $j\in\{1,\ldots,n_r\}$, we define residual-model input features
$\mathbf z^{(j)}$.
We write $\mathbf z=\bigl((\mathbf z^{(1)})^\top,\ldots,      (\mathbf z^{(n_r)})^\top\bigr)^\top$
for their stacked collection, with the $j$-th residual component depending only on
$\mathbf z^{(j)}$.
With a state vector $\mathbf{x}\in\mathbb{R}^{n_x}$ and
a control vector $\mathbf{u}\in\mathbb{R}^{n_u}$,
the ground-truth dynamics are
\begin{equation}
\dot{\mathbf{x}}
=
\mathbf f_{\mathrm{nom}}(\mathbf{x},\mathbf{u})
+
B_r\mathbf r^\star(\mathbf z).
\label{eq:continuous-residual-dynamics}
\end{equation}
Here, $\mathbf f_{\mathrm{nom}}$ denotes the known structured nominal dynamics,
$\mathbf r^\star(\mathbf z)\in\mathbb{R}^{n_r}$ denotes the unknown
ground-truth residual, and
$B_r\in\mathbb{R}^{n_x\times n_r}$ is the residual injection matrix.
Known context variables may additionally index separate residual models.
Such indices are suppressed in this section and made explicit in Sec.~\ref{subsec:sim_env_resi_model}.
Using a residual model $\mathbf r$ for prediction, explicit Euler
integration with step size $\Delta t$ gives
\begin{equation}
\mathbf{x}_{t+1}
=
\mathbf{x}_t+\Delta t\left[
 \mathbf f_{\mathrm{nom}}(\mathbf{x}_t,\mathbf{u}_t)
 +B_r\mathbf r(\mathbf z_t)
\right].
\label{eq:discrete-residual-dynamics}
\end{equation}

For each learned residual component $j$, we model the residual using an
independent scalar GP \cite{williams2006gaussian},
\begin{equation}
r_j(\mathbf z^{(j)})
\sim
\mathcal{GP}\!\left(
0,\kappa_j(\mathbf z^{(j)},\mathbf z^{\prime(j)})
\right).
\end{equation}
Using separate scalar GPs reduces the computational cost of online posterior
updates and rollout evaluation.
For each state transition, the residual observation is given by
\begin{equation}
y_{j,t}
=
r_j^\star(\mathbf z_t^{(j)})+\nu_{j,t},
\qquad
\nu_{j,t}\sim\mathcal N(0,\sigma_{n,j}^2),
\end{equation}
which gives the GP training pair
$(\mathbf z_t^{(j)},y_{j,t})$.
Given the accumulated training data, let
$\mu_j(\mathbf z^{(j)})$ and
$\Sigma_j(\mathbf z^{(j)},\mathbf z'^{(j)})$
denote the standard GP posterior mean and covariance, respectively.
The posterior mean is used for predictive rollouts, whereas the
posterior covariance is used by ToIA.
Joint residual models can also be used when the required predictive
covariances are available, although they may increase the cost of online
updates and covariance evaluation.

\begin{figure*}[ht]
  \centering
  \includegraphics[width=1.0 \hsize]{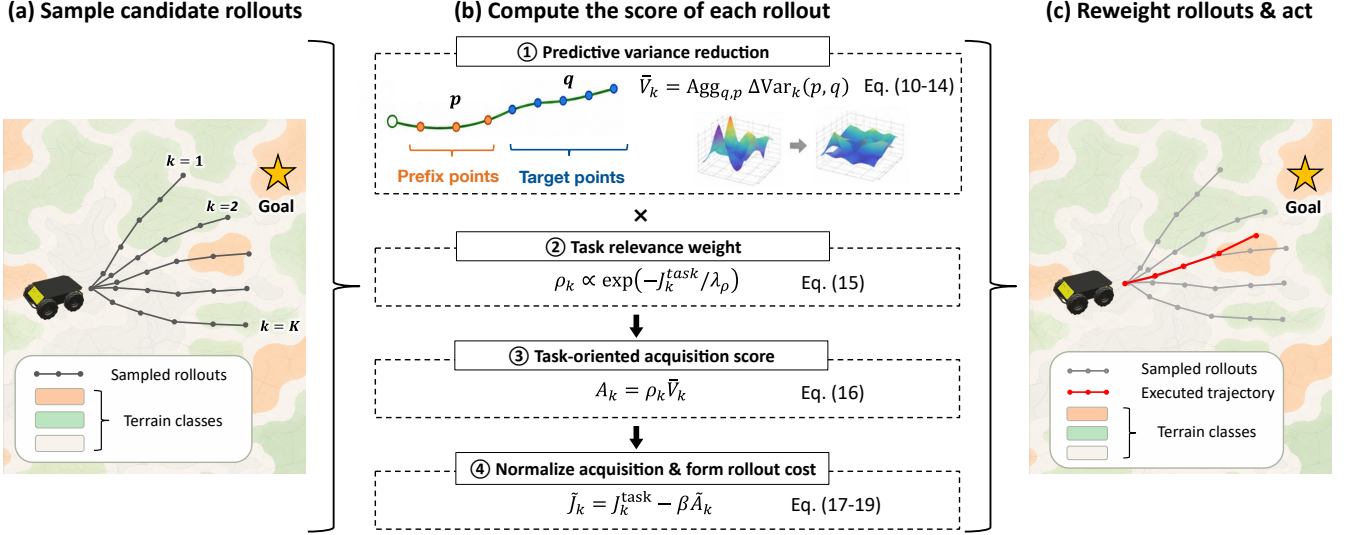}
  \caption{Overview of the proposed control scheme. Control sequences are first sampled and rolled out. Predictive variance reduction at later queries is then computed and combined with task relevance to score each rollout, and the next control action is obtained by reweighting the rollouts.}
  \label{fig:method-schematic}
\end{figure*}

\subsection{Receding-Horizon Control}
\label{subsec:receding_horizon}

At closed-loop time $t$, MPPI~\cite{williams_mppi_2017icra,williams2017model}
samples $K$ perturbed control sequences, each containing
$H$ control inputs, around a nominal sequence $\bar{\mathbf U}
=\{\bar{\mathbf u}_h\}_{h=0}^{H-1}$.
Here, $k\in\{1,\ldots,K\}$ indexes a sampled control sequence and
$h\in\{0,\ldots,H-1\}$ indexes the prediction step
within that sequence. The sampled controls are
\begin{equation}
\mathbf u_{k,h}
=
\bar{\mathbf u}_h+\boldsymbol\epsilon_{k,h},
\qquad
\boldsymbol\epsilon_{k,h}\sim\mathcal N(\mathbf 0,\Sigma_u),
\end{equation}
where $\Sigma_u$ is the covariance of the control perturbations.

For each sampled control sequence
$\mathbf U_k=\{\mathbf u_{k,h}\}_{h=0}^{H-1}$,
let $\mathbf x_{k,0}=\mathbf x_t$ and
$\mathbf x_{k,h+1}$ denote the state predicted after applying
$\mathbf u_{k,h}$. The corresponding rollout is
$\mathbf X_k=\{\mathbf x_{k,h}\}_{h=0}^{H}$, and the task
cost is
\begin{equation}
J_k^{\mathrm{task}}
=
\sum_{h=0}^{H-1}
\ell(\mathbf x_{k,h+1},\mathbf u_{k,h})
+
\ell_T(\mathbf x_{k,H}).
\label{eq:task-cost}
\end{equation}
Here, $\ell$ and $\ell_T$ denote the stage and terminal task costs,
respectively. The stage cost may include goal progress, control effort,
collision or boundary penalties, and speed regulation.

Let
$J_{\min}^{\mathrm{task}}=\min_{k\in\{1,\ldots,K\}} J_k^{\mathrm{task}}$.
With the temperature $\lambda>0$ controlling the concentration of the rollout weights, the normalized importance weight of rollout $k$ is
\begin{equation}
\label{eq:mppi-weights}
w_k=
\frac{
\exp[-(J_k^{\mathrm{task}}-J_{\min}^{\mathrm{task}})/\lambda]
}{
\sum_{k'=1}^{K}
\exp[-(J_{k'}^{\mathrm{task}}-J_{\min}^{\mathrm{task}})/\lambda]
}.
\end{equation}
The control applied at closed-loop time $t$ is then
\begin{equation}
\mathbf u_t
=
\bar{\mathbf u}_0+
\sum_{k=1}^{K}w_k\boldsymbol\epsilon_{k,0}.
\end{equation}
If a rollout enters the goal region before its terminal state, its task-cost accumulation is truncated at first entry, as in \cite{cai2024evora}.

\section{ToIA}

\label{sec:toia}
ToIA evaluates each sampled rollout separately. 
It estimates how much an observation obtained early in the rollout would reduce predictive uncertainty at later model queries on the same rollout.
ToIA combines this predictive value with the rollout's relevance to the current task.
Fig.~\ref{fig:method-schematic}
provides an overview of this procedure.

\subsection{Early Observations and Future Target Queries}
\label{subsec:prefix-target-points}

For each rollout $k$, let $\boldsymbol z_{k,h}^{(j)}$ denote the residual-model input at prediction step $h$, computed from the predicted state, control, and any known environmental features.
We select an early set of prospective observation indices and a later
set of target-query indices,
\[
\mathcal P=\{p_1<\cdots<p_{N_P}\},
\quad
\mathcal T=\{q_1<\cdots<q_{N_T}\},
\quad
p_{N_P}<q_1 .
\]
A prefix index $p\in\mathcal P$ denotes a step at which a residual observation could be obtained if rollout $k$ were executed, whereas $q\in\mathcal T$ denotes a later model query at which the value of that observation is assessed.
Rather than using a fixed global target set, ToIA defines target queries along the same rollout.
The score therefore quantifies how much an observation at a prefix point would reduce uncertainty at model evaluations that this rollout is predicted to encounter later.
These prefix points are prospective only; after scoring and reweighting, receding-horizon MPPI still applies only the first control action.

Since information after task completion is irrelevant to the current goal-reaching task, if rollout $k$ first enters the goal at $h_k^g$, post-goal queries are removed as
\begin{equation}
\mathcal T_k^g
=
\{q\in\mathcal T:q<h_k^g\}.
\label{eq:goal-aware-target}
\end{equation}
If the rollout does not enter the goal within the horizon, we set $\mathcal T_k^g=\mathcal T$.

\subsection{Predictive Variance Reduction}
For residual component $j$, consider a prospective observation at
prefix step $p$ and a later target query at step $q$.
Along rollout $k$, their posterior covariance is
\begin{equation}
\Sigma_{k,j}(q,p)
:=
\Sigma_j\!\left(
\boldsymbol z_{k,q}^{(j)},
\boldsymbol z_{k,p}^{(j)}
\right).
\end{equation}
Conditioning on an observation at prefix point $p$ with observation-noise
variance $\sigma_{n,j}^2$ reduces the posterior variance at target
$q$ by
\begin{equation}
\Delta\operatorname{Var}_{k,j}(p,q)
=
\frac{\Sigma_{k,j}(q,p)^2}
{\Sigma_{k,j}(p,p)+\sigma_{n,j}^2}.
\label{eq:variance-reduction}
\end{equation}
This reduction depends only on the current posterior covariance and observation-noise variance, not on the realized observation value.
Hence the information score can be evaluated without sampling a prospective measurement.

Let $\mathcal C_{\mathrm{learn}}$ denote the set of learned residual
components. We aggregate the component-wise reductions as
\begin{equation}
\Delta\mathrm{Var}_k(p,q)
=
\sum_{j\in\mathcal C_{\mathrm{learn}}}
\Delta\mathrm{Var}_{k,j}(p,q).
\label{eq:channel-aggregation}
\end{equation}

Let $\eta_{k,q}\ge 0$ denote the weight assigned to target query $q$.
For each prefix point $p$, we first combine the variance reductions over
the future target queries:
\begin{equation}
V_{k,p}
=
\sum_{q\in\mathcal T_k^g}
\eta_{k,q}\,
\Delta\operatorname{Var}_{k}(p,q).
\label{eq:prefix-target-value}
\end{equation}
We use log-sum-exp (LSE) with temperature $\tau$ to aggregate the prefix scores,
where a smaller $\tau$ emphasizes the largest score and a larger $\tau$ leads to more equal contributions among prefix points:
\begin{equation}
\bar V_k
=
\tau
\log
\sum_{p\in\mathcal P}
\exp\!\left(\frac{V_{k,p}}{\tau}\right),
\qquad \tau>0.
\label{eq:general-acquisition}
\end{equation}
If $\mathcal T_k^g=\emptyset$,
no acquisition score is applied and
the rollout retains $J_k^{\rm task}$.

\subsection{Task-Relevance Weight and Rollout Cost}
\label{subsec:task-rel-weight-cand-cost}

We define a rollout-level task-relevance weight from the task costs before adding the acquisition term:
\begin{equation}
\rho_k
=
\frac{
\exp[-(J_k^{\mathrm{task}}-J_{\min}^{\mathrm{task}})/\lambda_\rho]
}{
\sum_{i=1}^{K}
\exp[-(J_i^{\mathrm{task}}-J_{\min}^{\mathrm{task}})/\lambda_\rho]
}.
\label{eq:rho}
\end{equation}
The resulting acquisition score is
\begin{equation}
A_k=\rho_k\bar V_k.
\label{eq:acquisition-score}
\end{equation}
Only rollouts with at least one valid target query are included in the per-command standardization.
Let $\mathcal K_{\rm act}=\{k:\mathcal T_k^g\neq\emptyset\}$,
$K_{\rm act}=|\mathcal K_{\rm act}|$,
\begin{equation}
\mu_A
=
\frac{1}{K_{\rm act}}
\sum_{k\in\mathcal K_{\rm act}}
A_k,\quad
\sigma_A=
\sqrt{
\frac{1}{K_{\rm act}}
\sum_{k\in\mathcal K_{\rm act}}
(A_k-\mu_A)^2
}.
\end{equation}
If $\mathcal K_{\rm act}=\emptyset$, the acquisition term is skipped for that command.
Per-command standardization and clipping yield
\begin{equation}
\widetilde A_k
=
\operatorname{clip}\!\left(
\frac{A_k-\mu_A}{\sigma_A+\varepsilon_\sigma},
-c_A,c_A
\right).
\label{eq:normalized-acquisition}
\end{equation}
Here, $\varepsilon_\sigma$ is a numerical stabilizer for near-zero
rollout-score variance, and $c_A>0$ denotes the clipping threshold.

With the weighting coefficient $\beta$, the  rollout cost is
\begin{equation}
\widetilde J_k
=
J_k^{\mathrm{task}}
-
\beta\widetilde A_k.
\label{eq:toia-cost}
\end{equation}
The final MPPI weights are computed using~\eqref{eq:mppi-weights} and $\lambda$ with $J_k^{\mathrm{task}}$ replaced by $\widetilde J_k$.
The temperature $\lambda_\rho$ therefore controls only the
task-relevance weights in~\eqref{eq:rho}.

\subsection{Computational Complexity}
\label{subsec:computational_complexity}
Let $N$ be the number of retained observations per scalar GP model,
$C_\ell=|\mathcal C_{\mathrm{learn}}|$ the number of learned residual components,
$C_c$ the number of context classes when separate context-conditioned GP models are used ($C_c=1$ otherwise),
$K$ the number of sampled rollouts,
$H$ the rollout horizon, and
$N_P=|\mathcal P|$, $N_T=|\mathcal T|$.
Covariance evaluation between prefix points and target queries scales in the worst case as
\begin{equation}
O\!\left(
K C_\ell
\left[
N^2(N_P+N_T)+N N_PN_T
\right]
\right),
\end{equation}
where the \(N^2\) terms arise from triangular solves using cached Cholesky factors and the \(NN_PN_T\) term from covariance contraction.
Recomputing the Cholesky factors for all GP models costs
$O(C_\ell C_c N^3)$; rollout propagation remains sequential,
while target-query covariance evaluation can be parallelized.

\section{Experimental Design}
\label{sec:experimental_design}

We instantiate the proposed framework in a simulated off-road navigation setting with terrain-dependent residual dynamics. Height, slope, and discrete terrain-class maps are generated using a terrain-generation pipeline based on BenchNav \cite{endo2024benchnav}.
The experiments are designed to separate the effects of online residual learning and information acquisition using the same simulated plant, GP, and MPPI settings, while reserving independent terrain maps for held-out evaluation.

\subsection{Simulation Environment and Residual Model}
\label{subsec:sim_env_resi_model}

The off-road vehicle is represented by a minimal planar model.
The state contains planar pose and body-frame velocities, while the inputs directly command longitudinal and yaw acceleration.
The nominal model retains the body-frame kinematic coupling but omits terrain-dependent lateral and yaw-acceleration effects, which are represented by the residual model.

We define the state and control vectors as
\begin{equation}
\mathbf{x} =[p_x,p_y,\psi,v_x,v_y,\omega]^\top,
\qquad
\mathbf{u}=[a_{\mathrm{cmd}},\dot\omega_{\mathrm{cmd}}]^\top,
\end{equation}
where $(p_x,p_y)$, $\psi$, $(v_x,v_y)$, and $\omega$ denote inertial planar position, yaw, body-frame velocities, and yaw rate, respectively,
while $a_\text{cmd}$ and $\dot{\omega}_\text{cmd}$ denote longitudinal- and yaw-acceleration commands.

The controller is given the terrain-class and slope maps over the planning region.
At each predicted position, let $c_t\in\{1,2,3\}$ denote the terrain class obtained from the map.
The simulated dynamics are
\begin{equation}
\dot{\mathbf{x}}
=
\begin{bmatrix}
 v_x\cos\psi-v_y\sin\psi\\
 v_x\sin\psi+v_y\cos\psi\\
 \omega\\
 a_{\mathrm{cmd}}\\
 -v_x\omega\\
 \dot\omega_{\mathrm{cmd}}
\end{bmatrix}
+
B_r\boldsymbol r(\boldsymbol z_t,c_t),
\label{eq:nominal-residual-dynamics}
\end{equation}
where the first term defines the nominal dynamics $\mathbf f_{\mathrm{nom}}$, while the second term injects the residual, with $B_r=\left[ 0_{4 \times 2}; I_2 \right]$.
For an executed transition, the supervised residual label is the difference between the observed and nominally predicted acceleration in the corresponding residual channel, with the latter obtained from \eqref{eq:nominal-residual-dynamics} using $\mathbf r=\mathbf 0$.
In this study,
\begin{equation}
\mathbf r^\star =
[\Delta a_y^\star,\Delta\dot\omega^\star]^\top .
\end{equation}
The simulator and residual labels are noise-free; the observation-noise variance appears only in the GP likelihood model.

The controller also queries the local lateral slope $\gamma_y$ and defines the turning feature $\zeta=\max(v_x,0)\omega$.
The simulated residual dynamics are then
\begin{align}
\Delta a_y^{\star}
&=\operatorname{clip}\!\left(
-m_c k_s g\sin\gamma_y
-k_{\mathrm{turn}}\tanh\!\frac{\zeta}{z_{0,\mathrm{turn}}}
\right.\notag\\[-0.2em]
&\hspace{22mm}\left.\vphantom{\frac{\zeta}{z_{0,\mathrm{turn}}}}-d_yv_y,\ \pm a_{y,\max}\right),
\label{eq:true-lateral-residual}\\
\Delta\dot\omega^{\star}
&=\operatorname{clip}\!\left(
-k_{\mathrm{yaw}}\tanh\!\frac{\zeta}{z_{0,\mathrm{yaw}}}
-d_\omega\omega,\ \pm\dot\omega_{\max}\right),
\label{eq:true-yaw-residual}
\end{align}
where $g$ denotes gravitational acceleration.
The simulated plant uses
$k_s=0.16$, $k_{\mathrm{turn}}=1.43$, $k_{\mathrm{yaw}}=0.60$,
$z_{0,\mathrm{turn}}=0.0225$, $z_{0,\mathrm{yaw}}=0.225$,
$d_y=d_\omega=0.72$, $a_{y,\max}=2.5~\mathrm{m/s^2}$, and
$\dot\omega_{\max}=0.6~\mathrm{rad/s^2}$.
The class-dependent amplification factors are
\vspace{-2mm}
\begin{equation}
(m_1,m_2,m_3)=(1,2,3).
\label{eq:terrain-class-amplification}
\end{equation}
The terrain class $c\in\{1,2,3\}$ selects the corresponding multiplier $m_c$ in the cross-slope lateral residual in~\eqref{eq:true-lateral-residual}.
Hence class~1 represents the mildest terrain-dependent lateral residual, while classes~2 and~3 produce progressively larger residual effects.
The use of terrain-dependent residual magnitudes is motivated by the
importance of wheel--soil interaction and lateral forces in loose-soil
vehicle dynamics \cite{ishigami2007JFR}.
The turning, damping, and yaw-residual terms are shared across terrain classes.

The two learned residual channels use the channel-specific GP inputs
\begin{equation}
\boldsymbol z^{(\Delta a_y)}
=
[\gamma_y,\zeta,v_y]^\top,
\qquad
\boldsymbol z^{(\Delta \dot\omega)}
=
[\zeta,\omega]^\top.
\end{equation}
For the class-conditioned GP implementation, the terrain class serves as the context index.
Each of the three classes is assigned a separate GP model, so $C_c=3$ in these experiments.
Prefix and target queries from different terrain classes therefore have zero predictive covariance.

To evaluate performance across different terrain compositions, we use three held-out maps.
In the balanced map, the three terrain classes occupy comparable fractions of the map.
The class-1-heavy map is dominated by class~1, corresponding to a larger fraction of the mildest residual regime.
The class-2/3-heavy map instead contains a larger fraction of classes~2 and~3, exposing the controller more frequently to the larger-residual regimes.

\subsection{Task Objective and Method Settings}
\label{subsec:experimental_controller}
All methods solve the same goal-reaching navigation task with a goal-region radius of $0.5$ m.
We specialize~\eqref{eq:task-cost} to the experimental setting as follows. 
Let $\mathbf p_{k,h}$ denote the planar position component of
$\mathbf x_{k,h}$, $\mathbf p_g$ the goal position, and $\mathcal B$ the
valid map region:
\begin{align}
J_k^{\mathrm{task}}
={}&
w_g
\sum_{h=0}^{H-1}
\|\mathbf p_{k,h+1}-\mathbf p_g\|_2
+
w_u
\sum_{h=0}^{H-1}
\|\mathbf u_{k,h}\|_2^2
\nonumber\\
&+
w_b
\sum_{h=0}^{H-1}
\mathbf{1}\{
\mathbf p_{k,h+1}\notin\mathcal B
\}
+
w_T
\|\mathbf p_{k,H}-\mathbf p_g\|_2 .
\label{eq:experimental-task-cost}
\end{align}
We use $w_g=1$, $w_u=0.02$, $w_b=100$, and $w_T=10$
for running goal distance, control effort, boundary violation, and terminal goal distance, respectively.
We use the same first-goal-entry truncation as in Sec.~\ref{subsec:receding_horizon}.

The simulated plant uses the fixed ground-truth residual
$\mathbf{r^{\star}}$, regardless of the control method.
During controller rollouts, Nominal uses $\mathbf r=\mathbf{0}$, Oracle uses $\mathbf r=\mathbf r^\star$, and all GP-based methods use the current GP posterior means.
Passive GP (PG) performs online residual learning without an explicit information acquisition incentive. 
Pointwise uncertainty sampling (US) scores a rollout by the sum of
marginal GP posterior variances over its rollout queries and learned
channels, without accounting for covariance between query points or task
relevance.
Trajectory joint information gain (JIG) evaluates the selected rollout
queries jointly.
For residual component $j$ and terrain class $c$, let
$\boldsymbol{\Sigma}_{k,j,c}$ be the posterior covariance matrix over the
selected queries of rollout $k$ belonging to class $c$, and
$\sigma_{n,j,c}^2$ the observation-noise variance of the corresponding GP.

The JIG score is
\begin{equation}
A_k^{\mathrm{JIG}}
=
\sum_{j\in\mathcal C_{\mathrm{learn}}}
\sum_{c=1}^{C_c}
\frac{1}{2}
\log\det
\left(
I+\sigma_{n,j,c}^{-2}\boldsymbol{\Sigma}_{k,j,c}
\right).
\end{equation}
It therefore accounts for covariance among the selected queries while
remaining task-agnostic.
This baseline is motivated by prior work on trajectory-level active GP dynamics learning~\cite{buissonfenet2020jointIG}.
ToIA differs from both by valuing prospective prefix observations through their predictive variance reduction at future queries in the same rollout and weighting the resulting acquisition by the rollout's task relevance.
For US and JIG, the respective acquisition scores are standardized, clipped, and incorporated into the rollout cost analogously to \eqref{eq:normalized-acquisition}--\eqref{eq:toia-cost}.

Table~\ref{tab:setup_parameters} summarizes the common settings (panel (a)) and method-specific parameters (panel (b)).
All methods use the same plant, MPPI sampling budget, task objective, and learning intervals.
All GP-based methods share the same class-conditioned residual-GP representation and fixed GP hyperparameters.
For ToIA, the target query is assigned unit weight, $\eta_{k,q}=1$, and we use $\lambda_\rho=5$ and $c_A=3$.

\begin{table}[tb]
\caption{Common controller and method-specific configurations.}
\label{tab:setup_parameters}
\centering
\setlength{\tabcolsep}{4.2pt}
\vspace{-2mm}
\textbf{(a) Common numerical settings}
\vspace{0.6mm}
\begin{tabular}{@{}l c l c@{}}
\toprule
Parameter & Value & Parameter & Value \\
\midrule
$\Delta t$ & $0.05$ s 
& Max. control time & $10$ s (200 steps) \\
Goal radius & $0.5$ m & Success criterion & goal entry by $10$~s \\
Action bounds & $[-1,1]^2$
& Control-noise s.d. & {$(0.1,0.1)$} \\
{Sampled rollouts $K$ } & {$32$}
& Prediction horizon $H$ & $80$ \\
{Learned channels} & {$\Delta a_y,\Delta\dot\omega$}
& {GP kernel} & {RBF} \\
\bottomrule
\end{tabular}

\vspace{1.3mm}

\vspace{0.5mm}

\textbf{(b) Method-specific rollout and acquisition settings}
\vspace{0.6mm}

\begin{tabular}{@{}l c c c c @{}}
\toprule
Method & Scored query steps & Acquisition & $\beta$ & MPPI $\lambda$ \\
\midrule
Pointwise US & All rollout steps & Marginal variance & $0.5$ & $5$ \\

Trajectory JIG & $\{0,26,53,79\}$ & Log-det & $0.625$ & $5$ \\

ToIA (proposed) & $\{79\}^{\dagger}$  & LSE, $\tau=1$ & $5$ & $5$ \\
\bottomrule
\end{tabular}

\vspace{0.3mm}
{\footnotesize
$^{\dagger}$For ToIA, these are target-query steps; prefix steps are
$\{0,10,20,30\}$.
}

\vspace{-2mm}
\end{table}

\subsection{Tuning and Held-Out Data Split}

The tuning and validation maps are distinct.
GP hyperparameters were first calibrated using prequential predictive likelihood, without using task-success or acquisition metrics, and were then fixed and shared by all GP-based methods.
A separate 60-run tuning set was used to select the ToIA prefix and target steps, $\beta$, and $\lambda_\rho$, and the JIG query steps and acquisition gains.
Each method's acquisition gain was tuned independently on this same set, and all settings were frozen before validation.

\begin{table*}[t]
\caption{Held-out closed-loop performance on the three Held-Out validation maps.}
\label{tab:pool_v_main_results}
\centering
\vspace{-2mm}
\setlength{\tabcolsep}{3.0pt}
\begin{tabular}{@{}c l c c c c c c@{}}
\toprule
Learning interval
& Method
& Success
& R / H vs. PG
& Median $\Delta$ cmd. [$n$]
& F / T / S
& Ctrl. p50 / p95 [ms]
& \shortstack{95th pct. run p95 [ms]} \\
\midrule

\multirow{6}{*}{$(1,5)$}
& Nominal
& 111 / 135 (82.2\,\%)
& 18 / 5
& $-1$ [93]
& 50 / 19 / 24
& 21.03 / 21.42
& 23.80 \\

& Oracle
& 135 / 135 (100\,\%)
& 37 / 0
& $-8.5$ [98]
& 92 / 6 / 0
& 97.55 / 98.43
& 110.43 \\

\cmidrule(l){2-8}

& Passive GP
& 98 / 135 (72.6\,\%)
& --
& --
& --
& 20.84 / 24.54
& 25.82 \\

& Pointwise US
& 92 / 135 (68.1\,\%)
& 9 / 15
& {0 [83]}
& 29 / 18 / 36
& 27.36 / 31.42
& 34.28 \\

& Trajectory JIG
& 88 / 135 (65.2\,\%)
& 8 / 18
& {0 [80]}
& 30 / 14 / 36
& 35.67 / 39.80
& 43.98 \\

& ToIA (proposed)
& \textbf{124 / 135 (91.9\,\%)}
& \textbf{33 / 7}
& {0 [91]}
& {40 / 20 / 31}
& 31.99 / 39.13
& 44.39 \\

\midrule

\multirow{6}{*}{$(10,10)$}
& Nominal
& 111 / 135 (82.2\,\%)
& 23 / 4
& 0 [88]
& 39 / 32 / 17
& 21.05 / 21.45
& 24.01 \\

& Oracle
& 135 / 135 (100\,\%)
& 43 / 0
& $-7$ [92]
& 90 / 2 / 0
& 97.62 / 98.67
& 110.92 \\

\cmidrule(l){2-8}

& Passive GP
& 92 / 135 (68.1\,\%)
& --
& --
& --
& 17.66 / 22.61
& 23.72 \\

& Pointwise US
& 92 / 135 (68.1\,\%)
& 11 / 11
& {0 [81]}
& 24 / 19 / 38
& 23.69 / 28.38
& 30.24 \\

& Trajectory JIG
& 93 / 135 (68.9\,\%)
& 18 / 17
& $+1$ [75]
& 21 / 16 / 38
& 31.45 / 36.04
& 39.46 \\

& ToIA (proposed)
& \textbf{129 / 135 (95.6\,\%)}
& \textbf{38 / 1}
& {0 [91]}
& {35 / 15 / 41}
& 26.90 / 34.54
& 39.68 \\

\bottomrule
\end{tabular}
\vspace{-2mm}
\end{table*}

\subsection{Evaluation Protocol and Metrics}

Observed transitions are added to the GP training set every
$N_{\mathrm{obs}}$ commands.
GP hyperparameters are fixed during each run, while posterior Cholesky factors are updated every $N_{\mathrm{fit}}$ commands.
We evaluate two learning intervals:
a dense setting, $(N_{\mathrm{obs}},N_{\mathrm{fit}})=(1,5)$, and a sparse setting, $(10,10)$.

The held-out evaluation contains 15 start--goal pairs on each of three validation maps and three control seeds for each pair. This gives 45 runs per map and 135 runs in total for each method and learning interval.
Goal success is defined as entering a goal region with a radius of $0.5$ m within 200 control commands ($10$ s).
Paired comparisons use Passive GP (PG) as the reference. A rescue (R) is a run in which PG fails but the compared method succeeds; a harm (H) is the converse.
Completion time is compared only on runs where both methods succeed.
We compare the number of control commands required to first enter the goal region.
Command-count differences are computed as the compared method
minus Passive GP, so negative values indicate earlier goal entry.
F/T/S reports how often the compared method is faster, tied, or slower on these common-success runs; $n$ in Table~\ref{tab:pool_v_main_results} is the number of such runs.
The 95\,\% confidence intervals for success-rate differences are obtained by resampling the 45 start--goal scenarios while keeping the three control seeds of each scenario together.

Controller runtime is measured on an NVIDIA RTX 2080 Ti.
For each run, we compute the p50 and p95 end-to-end command latencies over all
executed control updates.
Table~\ref{tab:pool_v_main_results} reports the median of each statistic across the 135 runs and the 95th percentile across the run-level p95 latencies.

\section{Results}
\label{sec:results}

\subsection{Held-Out Task Performance}

\begin{figure}[t]
  \centering
  \includegraphics[width=1.0 \hsize]{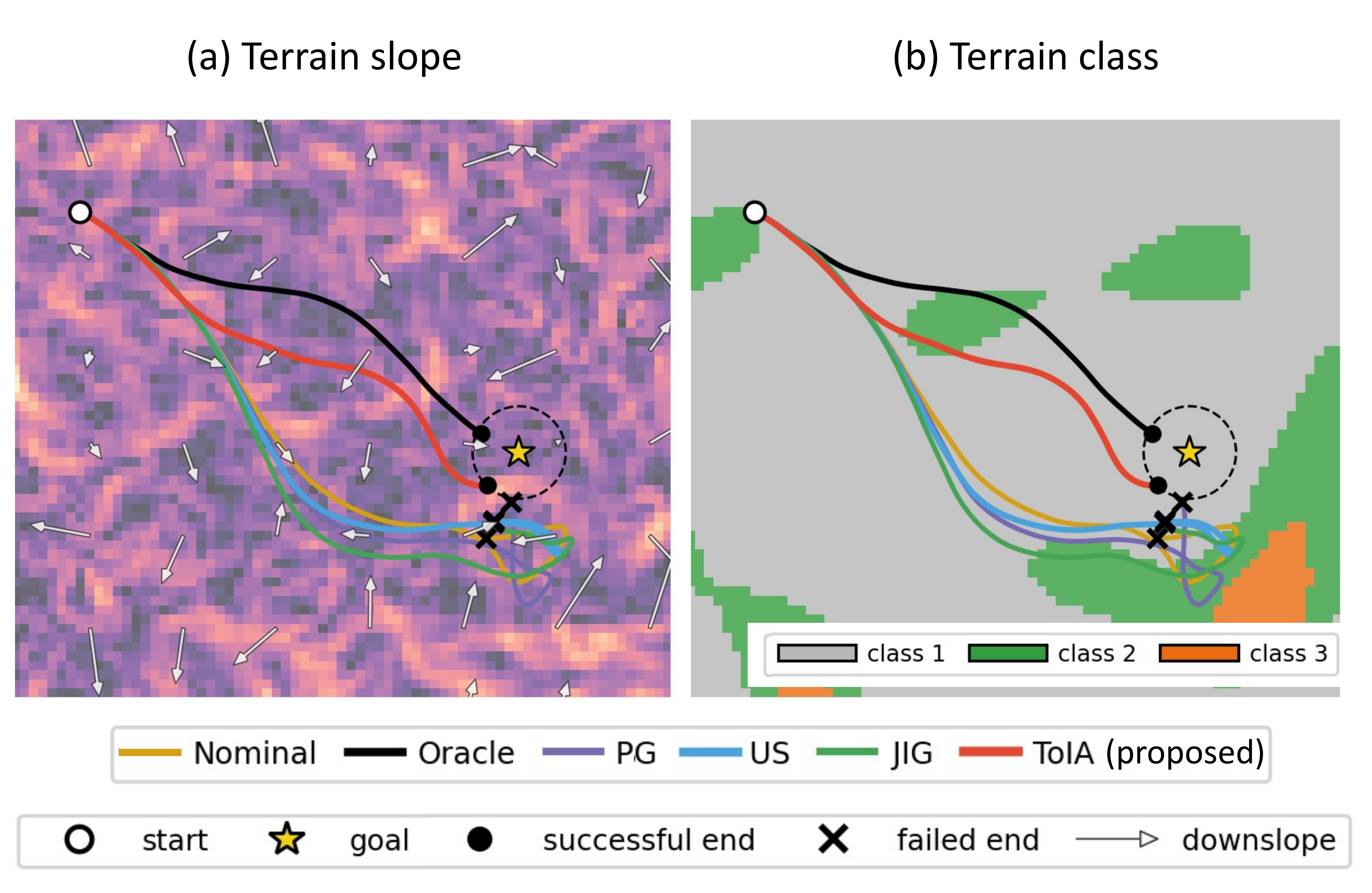}
  \vspace{-6mm}
  \caption{Example closed-loop trajectories on a held-out task with heterogeneous terrain. (a) Terrain slope with downslope vectors, whose direction indicates local descent and whose length indicates relative slope magnitude. (b) Corresponding terrain-class map. The dashed circle indicates the goal region.
  }
  \label{fig:example-trajectories}
  \vspace{-2mm}
\end{figure}

Fig.~\ref{fig:example-trajectories} shows trajectories on a held-out heterogeneous-terrain task, illustrating how the learned residual model and acquisition strategy affect the executed route.
Table~\ref{tab:pool_v_main_results} reports held-out task performance and paired comparisons against Passive GP.
On the held-out terrain maps, ToIA
achieved 91.9\,\% and 95.6\,\% goal success under the dense and sparse settings, respectively, outperforming Passive GP and both task-agnostic active-learning baselines.
Pointwise US and trajectory JIG did not provide a consistent improvement over Passive GP.
Table~\ref{tab:pool_v_main_results} also reports the end-to-end controller latency.
For ToIA, the 95th percentile across run-level p95 latencies was 44.39 ms under the dense setting and 39.68 ms under the sparse setting, below the 50-ms control period.
The reported latencies measure the complete controller update rather than the acquisition computation alone; differences in trajectories and episode lengths may therefore also affect the observed runtime distributions.

\begin{figure}[t]
  \centering
    \includegraphics[width=1.01 \hsize]{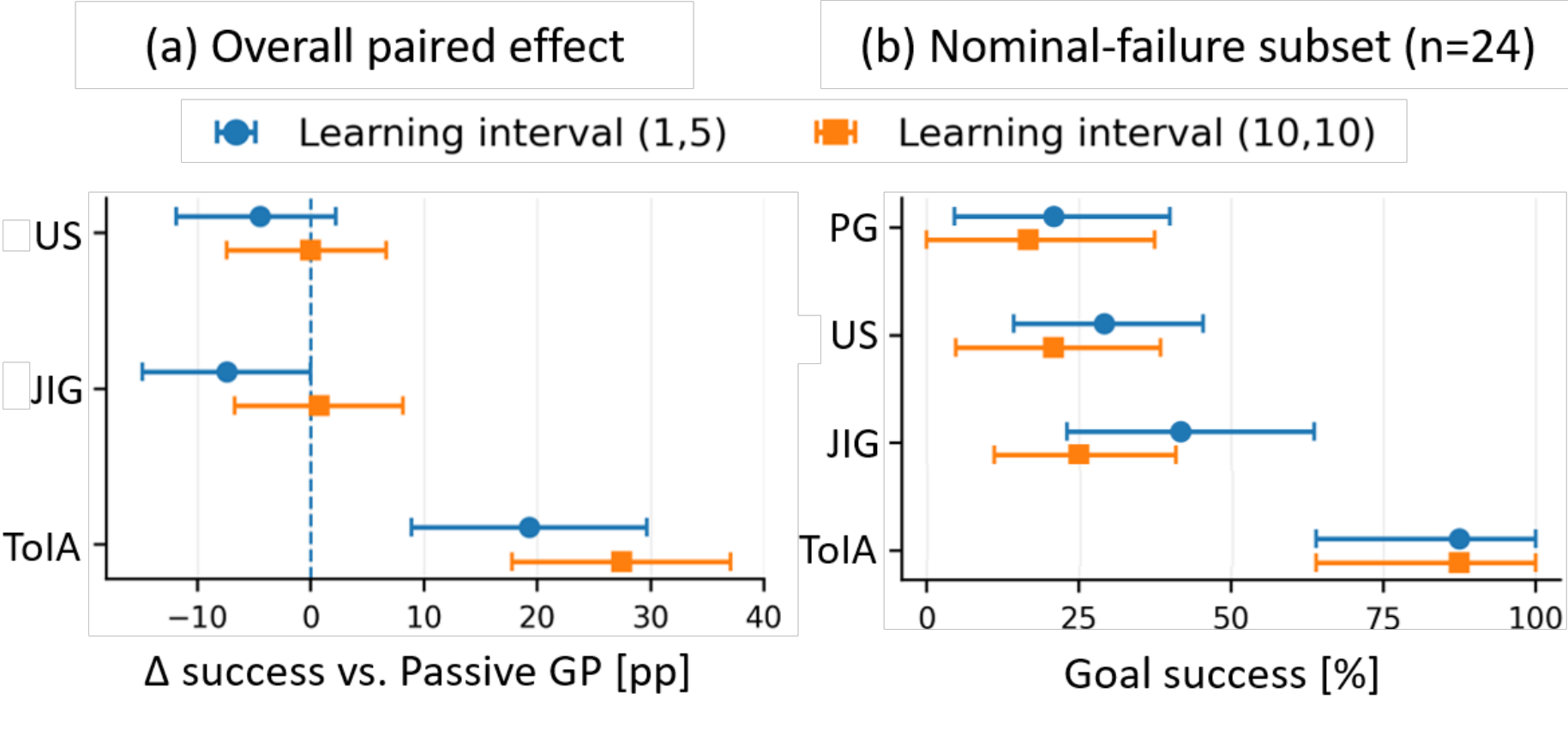}
    \vspace{-7mm}
  \caption{Held-out paired success effects.
        (a) Paired change in goal success rate relative to Passive GP over the 135 matched runs per learning interval.
        Whiskers denote 95\,\% confidence intervals obtained by resampling start--goal scenario clusters within each of the three fixed validation maps, retaining the three control seeds of each scenario.
        (b) Goal success on the fixed 24 runs for which the non-learning
        Nominal controller failed. This subset was defined independently of
        the active-method outcomes.}
  \label{fig:paired-outcomes}
  \vspace{-1mm}
\end{figure}

Over all 135 held-out runs, ToIA improved goal success relative to
Passive GP by 19.3 and 27.4 percentage points under the dense and sparse settings, respectively, with the 95\,\% confidence intervals remaining above zero (Fig.~\ref{fig:paired-outcomes}(a)).
This paired comparison further shows that ToIA rescued 33 Passive GP failures while harming 7 Passive GP successes under the dense setting.
Under the sparse setting, it produced 38 rescues and only 1 harm. Relative to JIG, the corresponding rescue/harm counts were 42/6 and 38/2. 
On the fixed Nominal-failure subset in Fig.~\ref{fig:paired-outcomes}(b), ToIA succeeded in 21/24 runs under both learning intervals, whereas the learning baselines achieved at most 10/24.
Among the remaining learning-based methods, Trajectory JIG achieved the highest success count on this subset.
Completion-time effects were mixed, with a median difference of 0 commands for ToIA under both intervals.

\subsection{Performance Across Terrain Maps}
Table~\ref{tab:terrain_family_results} reports goal success separately for the three held-out terrain maps.
ToIA achieved the highest success count in all six learning interval--terrain combinations.
Its improvement over Passive GP was largest on the class-2/3-heavy terrain, where ToIA achieved 11 more successes under dense learning and 15 more under sparse learning.

\begin{table}[t]
\caption{Goal success on the three held-out validation maps.}
\label{tab:terrain_family_results}
\centering
\vspace{-2mm}
\setlength{\tabcolsep}{5.0pt}
\begin{tabular}{@{}c l c c c@{}}
\toprule
Interval & Method & Balanced & Class-1-heavy & Class-2/3-heavy \\
\midrule

\multirow{4}{*}{$(1,5)$} & Passive GP & 37 / 45 & 33 / 45 & 28 / 45 \\
& Pointwise US & 34 / 45 & 33 / 45 & 25 / 45 \\
& Trajectory JIG & 34 / 45 & 30 / 45 & 24 / 45 \\
& ToIA (proposed) & \textbf{44 / 45} & \textbf{41 / 45} & \textbf{39 / 45} \\

\midrule

\multirow{4}{*}{$(10,10)$} & Passive GP & 35 / 45 & 31 / 45 & 26 / 45 \\
& Pointwise US & 35 / 45 & 33 / 45 & 24 / 45 \\
& Trajectory JIG & 34 / 45 & 34 / 45 & 25 / 45 \\
& ToIA (proposed) & \textbf{44 / 45} & \textbf{44 / 45} & \textbf{41 / 45} \\

\bottomrule
\end{tabular}
\vspace{-2mm}
\end{table}

\subsection{Ablation and Mechanism Analysis}

\begin{figure}[t]
  \centering
    \includegraphics[width=0.99 \hsize]{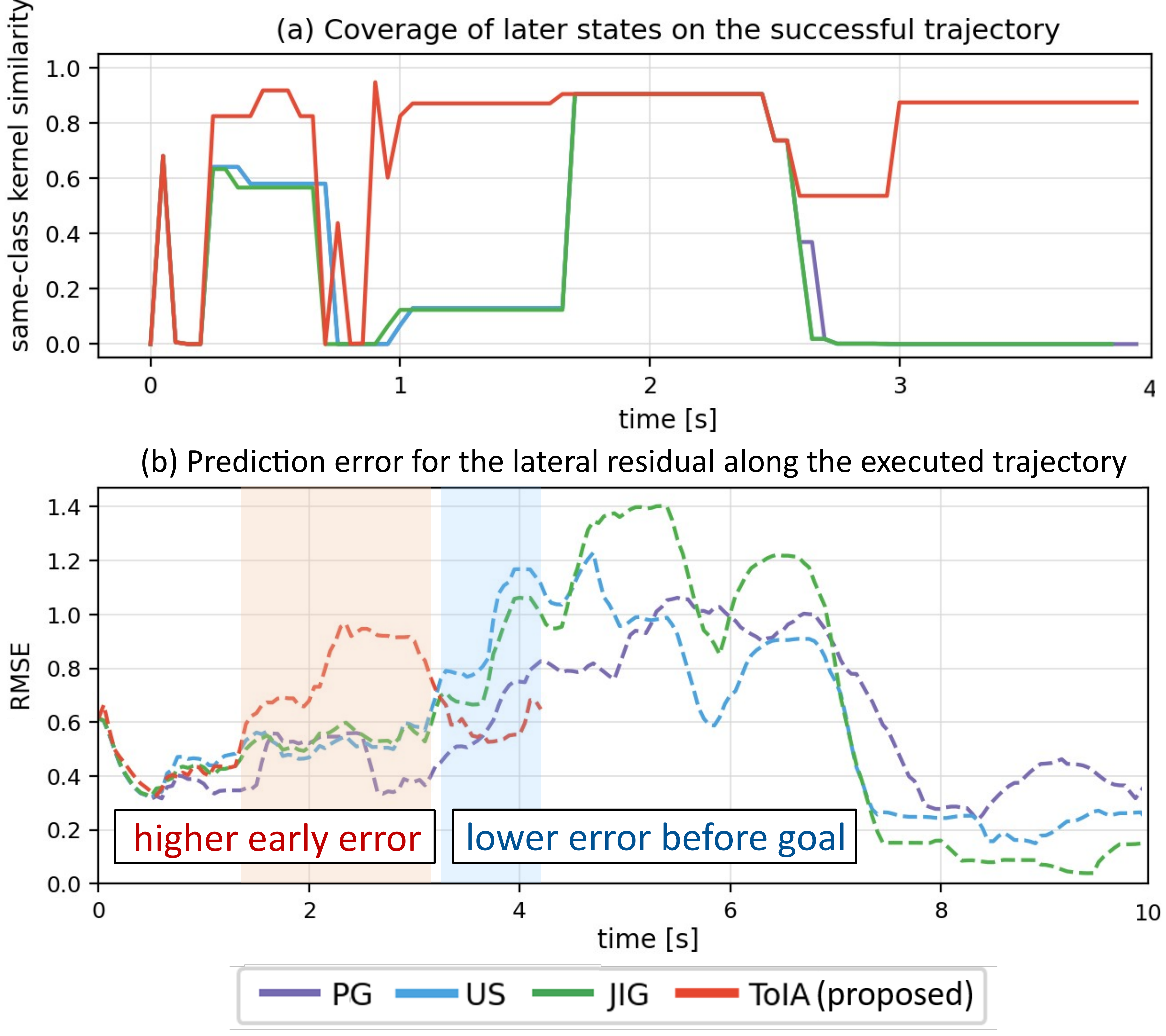}
    \vspace{-6mm}
  \caption{Analysis of a dense-learning case. (a) Coverage of later states on the matched successful Oracle trajectory, measured by the maximum same-class GP-kernel similarity between the current training data and the trajectory states. (b) RMSE of the pre-update lateral-residual prediction, evaluated along each trajectory over a trailing 1 s window. Shading highlights the transition from higher early ToIA error to lower error before goal entry. Each curve ends when the corresponding run terminates (goal entry or timeout).}
  \label{fig:mechanism-analysis}
\end{figure}

\begin{table}[t]
\caption{Fixed-configuration ablation of task weighting $\rho$.}
\label{tab:pool_v_ablation}
\vspace{-2mm}
\centering
\setlength{\tabcolsep}{3.2pt}
\begin{tabular}{@{}c c c c@{}}
\toprule
Interval & Task weight $\rho$ & Success & R / H vs. $\rho$-off \\
\midrule

\multirow{2}{*}{$(1,5)$}
& Off ($\rho=1$) & 104 / 135 (77.0\,\%) & -- \\
& On (ToIA) & \textbf{124 / 135 (91.9\,\%)} & 28 / 8 \\

\midrule

\multirow{2}{*}{$(10,10)$}
& Off ($\rho=1$) & 91 / 135 (67.4\,\%) & -- \\
& On (ToIA) & \textbf{129 / 135 (95.6\,\%)} & 42 / 4 \\
\bottomrule
\end{tabular}
\vspace{-2mm}
\end{table}

Table~\ref{tab:pool_v_ablation} evaluates the contribution of task relevance by disabling only the task-relevance weight. Without this weight, success remained above the learning baselines under the dense setting but dropped to baseline levels under the sparse setting, indicating that task relevance becomes especially important when model updates are infrequent.

We further examine one dense-learning case.
Fig.~\ref{fig:mechanism-analysis}(a) quantifies training-data coverage using the normalized RBF-kernel similarity
$s(\mathbf z,\mathbf z_i)=\kappa_{\Delta a_y}(\mathbf z,\mathbf z_i)/\sigma^2_{f,\Delta a_y}$,
where $\sigma^2_{f,\Delta a_y}=\kappa_{\Delta a_y}(\mathbf z,\mathbf z)$ is the
signal variance of the $\Delta a_y$ kernel, so that $s\in(0,1]$.
At each step, we consider up to 15 states on the matched Oracle trajectory whose projections onto the start--goal axis lie at or beyond the vehicle's current projection.
The plotted value is the maximum same-class GP-kernel similarity between their queries and the current training data.
In this example, PG and the task-agnostic baselines exhibit low similarity to upcoming Oracle states over several portions of the trajectory, whereas ToIA maintains higher similarity.
Fig.~\ref{fig:mechanism-analysis}(b) shows that ToIA has larger lateral-residual prediction error early in the episode but lower error than the baselines during the later phase before goal entry.
Together, these observations are consistent with ToIA acquiring data that improve the model where it is subsequently needed for the task, rather than seeking information without considering downstream task.
This may also help explain why the learning-based control baselines did not outperform Nominal: when task-relevant regions remain poorly covered, inaccurate residual corrections can degrade prediction more than omitting the residual model altogether does.

\section{Discussion}
\label{sec:discussion}

The ablation and mechanism analysis indicate that predictive acquisition and task relevance play complementary roles, particularly when model updates are sparse.
The proposed acquisition is a computationally tractable surrogate for Bayesian value-of-information planning.
Gaussian conditioning provides exact predictive variance reduction for each prefix--target pair, while evaluating these quantities on the existing MPPI rollout batch avoids sampling future observations or re-optimizing control under hypothetical posterior updates.
Explicitly modeling how posterior updates alter subsequent control decisions is left for future work.

The present study uses synthetic residual dynamics and independent GP models for each terrain class.
Evaluation with more realistic vehicle dynamics and physical systems remains future work.

\section{Conclusion}
\label{sec:conclusion}

In this paper, we presented ToIA, which augments MPPI with task-weighted predictive information acquisition during online GP residual learning.
Across three held-out maps, ToIA achieved goal success rates of 91.9\,\% and 95.6\,\% under the dense and sparse learning intervals, improving over Passive GP by 19.3 and 27.4 percentage points, respectively.
On the fixed 24-run subset in which the Nominal controller failed, ToIA succeeded in 21 runs under both intervals, while task-agnostic active-learning baselines did not reproduce this improvement.
An ablation study showed that predictive acquisition without task weighting retained 77.0\,\% success under the (1,5) setting but only 67.4\,\% under the (10,10) setting, suggesting that predictive acquisition and task relevance play complementary roles across learning intervals.
These results show that task-oriented information acquisition can improve closed-loop control with online residual learning while supporting 20-Hz online execution on an NVIDIA RTX 2080 Ti.

%\section*{Acknowledgement}

\bibliographystyle{IEEEtran}
\bibliography{reference}

\end{document}